%% file: main.tex
\documentclass{article} %
\usepackage{iclr2023_conference,times}

\input{math_commands.tex}

\usepackage{hyperref}
\usepackage{url}
\usepackage{graphicx}
\usepackage{url,booktabs,multirow,caption,subcaption,bm}

\title{Solar Panel Mapping via \\ Oriented Object Detection}

\author{Conor Wallace, Isaac Corley \& Jonathan Lwowski \\
Zeitview\\
\texttt{\{firstname.lastname\}@zeitview.com} \\
}

\iclrfinalcopy %
\begin{document}

\maketitle

\input{sections/abstract.tex}
\input{sections/introduction.tex}
\input{sections/related.tex}
\input{sections/method.tex}
\input{sections/experiments.tex}
\input{sections/conclusion.tex}

\bibliography{references}
\bibliographystyle{iclr2023_conference}

\input{sections/appendix.tex}

\end{document}

%% file: math_commands.tex
\usepackage{amsmath,amsfonts,bm}

\def\eqref#1{equation~\ref{#1}}

\def\1{\bm{1}}

\DeclareMathAlphabet{\mathsfit}{\encodingdefault}{\sfdefault}{m}{sl}
\SetMathAlphabet{\mathsfit}{bold}{\encodingdefault}{\sfdefault}{bx}{n}

%% file: sections/abstract.tex
\begin{abstract}
Maintaining the integrity of solar power plants is a vital component in dealing with the current climate crisis. This process begins with analysts creating a detailed map of a plant with the coordinates of every solar panel, making it possible to quickly locate and mitigate potential faulty solar panels. However, this task is extremely tedious and is not scalable for the ever increasing capacity of solar power across the globe. Therefore, we propose an end-to-end deep learning framework for detecting individual solar panels using a rotated object detection architecture. We evaluate our approach on a diverse dataset of solar power plants collected from across the United States and report a mAP score of $83.3\%$.
\end{abstract}

%% file: sections/introduction.tex
\input{figures/solarnet_architecture.tex}

\section{Introduction}

The adoption of renewable energy resources is paramount to fighting the climate change crisis. Over the last decade renewable energy production has nearly quadrupled, 26\% of which is contributed by solar energy \citep{renewables}. To ensure that solar power plants are operating at maximum capacity, inspections utilizing remotely sensed imagery are important in identifying anomalies in damaged panels. However, with this rapid increase in solar production, it has become increasingly difficult to scale these inspections efficiently. The initial step in many inspection pipelines is to localize and georeference individual panels for downstream evaluation tasks. This can be prohibitively time-consuming for commercial sites, any of which can easily contain tens of thousands of solar panels. Therefore, this requires an automated approach for detecting and georeferencing individual solar panels from large scale aerial imagery. To be specific, we wish to create a model that directly predicts the coordinates of the vertices of each solar panel. We also require the approach to generalize to images of solar arrays in any orientation across a wide variety of landscapes and environments.

In this paper we introduce a rotated object detection framework for localizing individual solar panels with arbitrary orientation. We preprocess large-scale orthomosaics into patches which are then processed in batches. The predictions are stitched back together and projected from pixel coordinates to geospatial coordinates, creating an accurate mapping of individual panels in each site. To the best of our knowledge this is the first study to model solar panel detection as a rotated object detection problem which allows us to efficiently map individual solar panels in an end-to-end fashion. 

%% file: figures/solarnet_architecture.tex
\begin{figure}[ht!]
\centering
\includegraphics[width=\textwidth]{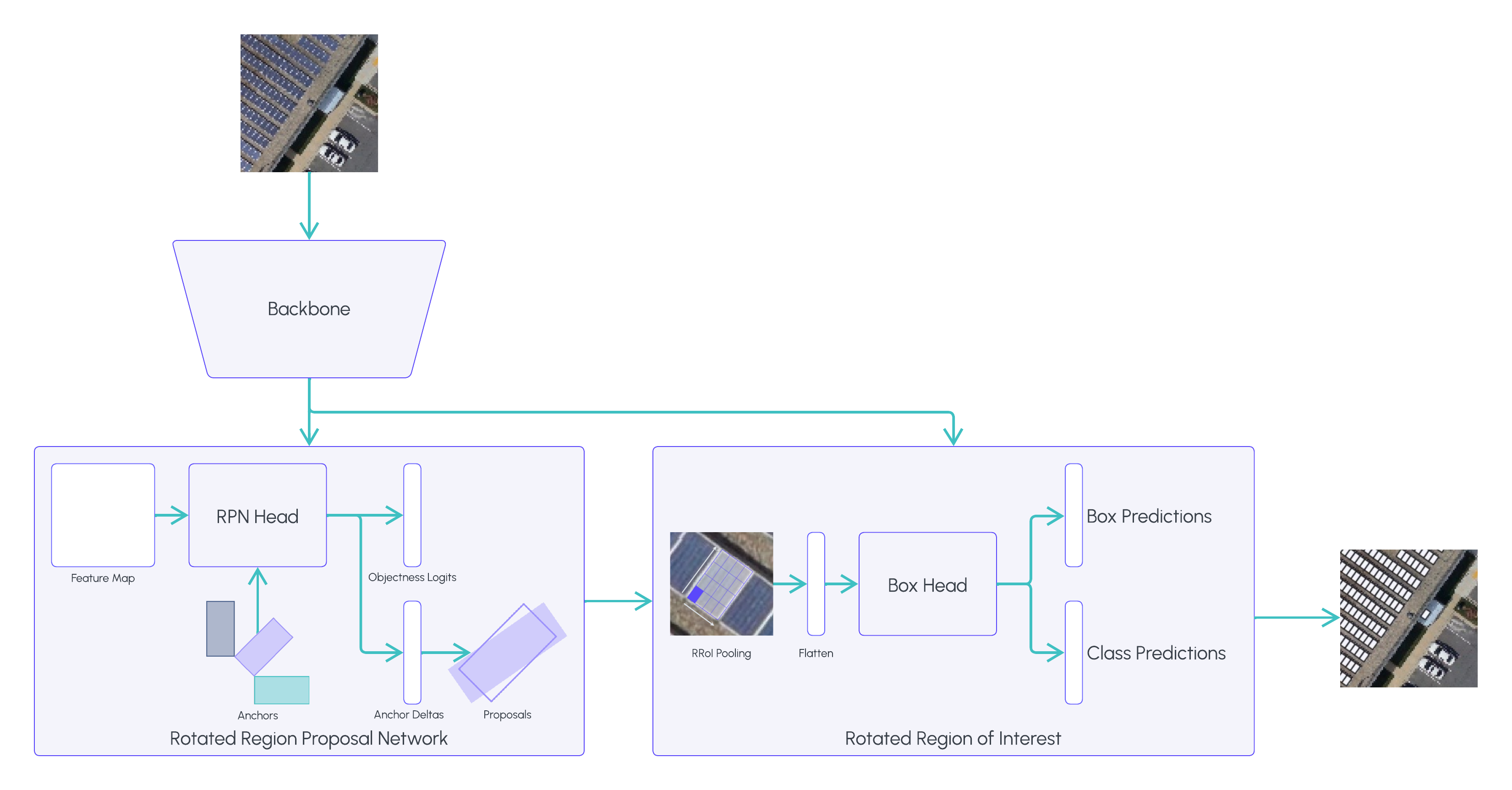}
\caption{Oriented Panel Detection Architecture.}
\label{fig:architecture}
\end{figure}

%% file: sections/related.tex
\section{Related Work}
Most recent works build algorithms for mapping solar panels from aerial or satellite imagery \citep{DeepSolar, SegNet, MujtabaAndWani, GolovkoEtAl} with the goal of estimating surface area as a proxy for energy capacity, location density, and other potential insights. These studies have focused on both small-scale residential solar sites \citep{CrossNets, SolarDK} as well as large-scale commercial sites \citep{DeepSolar}. Most approaches use a two-stage architecture for mapping solar panels from aerial imagery \citep{SolarNet, HyperionSolarNet}. The first stage consists of a convolutional neural network (CNN) classifier for predicting the probability of the given image containing solar panels or not. The second stage uses a segmentation model for segmenting out solar panels, contingent upon the first stage predicting the given image to contain solar panels. While these methods have shown promise in mapping the surface area of solar arrays across a diverse set of geographies, it remains nontrivial to extract the coordinates of individual solar panel vertices at scale. 

%% file: sections/method.tex
\section{Methodology}
In this section, we describe the proposed framework's architecture and the modeling of detection targets for solar panels.

\subsection{Rotated Bounding Boxes}
\label{section:bounding_boxes}
Our key desideratum is to accurately localize individual solar panels, regardless of the orientation of the panels in the image, and to use this localization to extract the coordinates of the vertices of each panel. Accordingly, we model the problem as an arbitrarily oriented object detection task where the ground truth annotations of the solar panels are represented as rotated bounding boxes. Each box is comprised of the 5-dimensional tuple $(x, y, w, h, \theta)$ where $(x, y)$ is the centroid pixel coordinate of the box, $w$ is the width defined as the shorter side of the box, $h$ is the height defined as the longer side of the box, and $\theta$ is the rotation angle in degrees required to obtain the rotated box. This bounding box formulation provides a compact representation which allows us to easily convert to the complete set of vertex coordinates $(x_1, y_1, x_2, y_2, x_3, y_3, x_4, y_4)$ using equation \ref{eq:vertices} in Appendix \ref{appendix:implementation}.

\subsection{Rotated Anchors}
Following R-CNN style object detection architectures, we define a set of proposal anchors as a strong basis for regression to ground truth rotated bounding boxes. In addition to scale and aspect ratio, we also provide rotation angle anchors. We specify seven different rotation anchors: $-90^{\circ}$, $-60^{\circ}$, $-30^{\circ}$, $0^{\circ}$, $30^{\circ}$, $60^{\circ}$, and $90^{\circ}$. We use standard anchors for scale and aspect ratio parameters which are defined as $[32, 64, 128, 256, 512]$ and [1:2, 1:1, 2:1], respectively. This produces a set of 105 anchors in total (7 rotation anchors, 5 scale anchors, and 3 aspect ratio anchors).

\subsection{Rotated Object Detection}
We adopt a general Faster R-CNN architecture \citep{FasterRCNN} with a CNN backbone, region proposal network (RPN), classification and box regression head, and a region of interest (RoI) pooling layer. We use the oriented object detection framework from \citep{RRPN} in which the authors present rotated RPN (RRPN) and rotated RoI (RRoI) modifications. The RRPN performs two functions: it produces rotated bounding box proposals from the specified box anchors defined in Section \ref{section:bounding_boxes} and regresses the box proposals to accurately localize panel instances for each feature map. From the RRPN, the proposals and feature maps are fed into the RRoI pooling layer which projects the arbitrarily oriented proposals onto each feature map and subsequently performs a max pooling operation. The pooled features from the RRoI layer are fed into a classification head and a box regression head which produce class labels, class scores, and box coordinates 
for each solar panel instance. The architecture can be seen in Figure. \ref{fig:architecture}.

%% file: sections/experiments.tex
\input{figures/results_plot.tex}

\section{Experiments}

\subsection{Dataset}

To train the model, we create a diverse dataset of aerial imagery from various locations across North America. Figure \ref{fig:usa-plot} in Appendix \ref{appendix:dataset} shows the exact locations where the data is collected. The dataset consists of 121 high-resolution stitched orthomosacis of large-scale solar farms captured using a fixed-wing aircraft with an RGB camera at approximately 2.5cm ground sample distance (GSD).

\subsection{Dataset Preprocessing}
We randomly split the 121 orthomosaics into train, validation, and test sets using an approximate split ratio of 80/10/10. We then preprocess each orthomosaic into 512x512 patches. Annotating large scale orthomosaics is time consuming and expensive. Additionally, since visual features of solar panels do not vary in comparison to natural images, we find it to be redundant to fully annotate and train on all patches from each site. Consequently, we randomly sample 10 unique foreground patches containing solar panels from each orthomosaic for annotating. We additionally sample 5 unique background patches from each orthomosaic to increase model precision on images of solar arrays with substantial portions of background information. We find that with proper augmentation, described in Appendix ~\ref{section:augmentation}, we achieve better results with a sampled dataset from a larger set of orthomosaics than we do with a smaller set of fully annotated orthomosaics. Finally, to test the efficacy of the proposed solar panel mapping framework, we create complete georeferenced maps of the 10 test orthomosaics containing the geospatial coordinates of the vertices of every solar panel. The spread of all three of our datasets is depicted in Table \ref{tab:dataset} in Appendix \ref{appendix:dataset}.

\subsection{Metrics}
To evaluate our model, we report mean average precision (mAP) and mean average recall (mAR) as our primary performance metrics. We also report mAP and mAR with an IoU threshold of 0.75, notated as $\text{mAP}_{75}$ and $\text{mAR}_{75}$, respectively. This is because we are interested in the performance of the model at high IoU thresholds as a proxy for evaluating how tight the localizations are.

\input{tables/results.tex}

\subsection{Results}
We fine-tune the model using several different combinations of ResNet backbone networks, all of which are pretrained on the ImageNet dataset \citep{imagenet_cvpr09}:

\begin{itemize}
    \item C4: A conv4 backbone with a conv5 head as described in the original Faster R-CNN paper \citep{FasterRCNN}.
    \item DC5: A conv5 backbone with dilations in the conv5 layer and fully connected layers in the box regression head \citep{DC}. 
    \item FPN: A feature pyramid network backbone with fully connected layers in the box regression head \citep{FPN}.
\end{itemize}

We run an ablation study to investigate the effects of the choice of backbone on model performance. We don’t run any hyperparameter tuning to keep things fair and we use the same hyperparameters for every model, with the exception that we reduce the batch size for deformable convnet backbones which require more memory during processing. The ResNet-50-FPN network comfortably achieves the best results on the validation set and comparable results on the test set.

\subsection{Discussion}
The growth in cumulative solar capacity has made it difficult to scale up the inspection and maintenance of these valuable assets. In 2022, we estimate that 3.5 GW of potential solar power was lost due to faulty solar panels. By automating the initial mapping phase of the inspection process, we estimate that analysts will have a 43\% increase in efficiency. This increase in analyst efficiency decreases the time that faulty solar panels are losing potential power. Finally, we estimate that our increase in inspection efficiency will lead to a potential retention of 1.4 GW of solar power.

%% file: figures/results_plot.tex
\begin{figure}[t!]
\centering
\includegraphics[width=\textwidth]{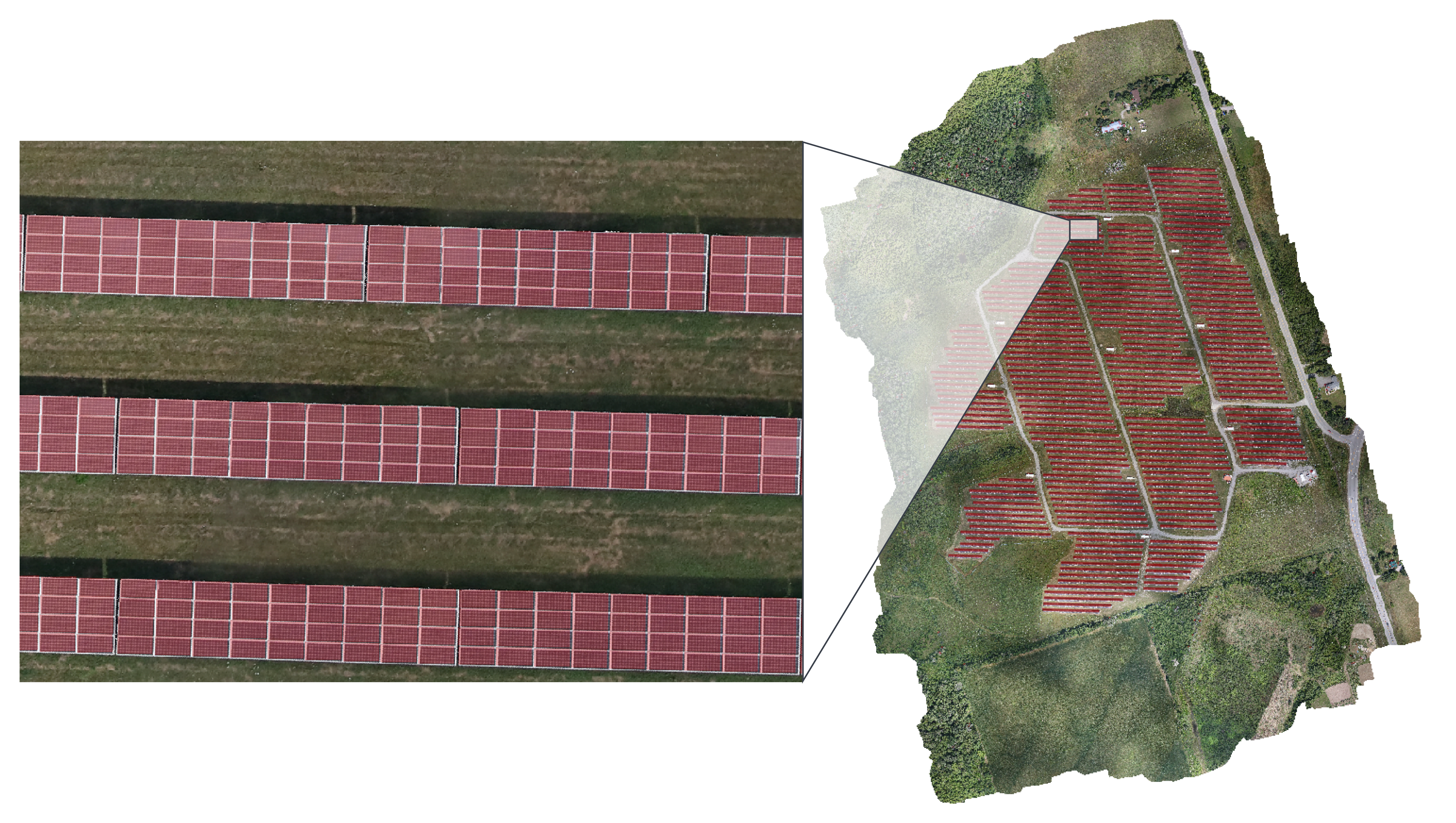}
\caption{Sample Test Set Ortho Panel Predictions.}
\label{fig:results}
\end{figure}

%% file: tables/results.tex
\begin{table}[t!]
\centering
\resizebox{\columnwidth}{!}{\begin{tabular}{@{}lcccccccc@{}}
\toprule
\multicolumn{1}{c}{} & 
\multicolumn{4}{c}{\textbf{Validation}} &
\multicolumn{4}{c}{\textbf{Test}} \\
\cmidrule(lr){2-5}\cmidrule(lr){6-9}
\multicolumn{1}{c}{\textbf{Backbone}} &
\textbf{AP} &
\textbf{AR} &
\textbf{$\text{AP}_{75}$} &
\textbf{$\text{AR}_{75}$} &
\textbf{AP} &
\textbf{AR} &
\textbf{$\text{AP}_{75}$} &
\textbf{$\text{AR}_{75}$} \\ 
\midrule
ResNet-50-C4         &80.9\%         &84.1\%          &92.5\%         &93.2\%         &84.8\%         &88.1\%          &97.9\%         &\textbf{98.4\%} \\
ResNet-50-DC5         &80.1\%         &84.1\%          &92.0\%         &93.2\%         &85.0\%         &88.3\%          &97.9\%         &98.3\% \\
ResNet-50-FPN         &\textbf{83.3\%}         &\textbf{88.4\%}          &\textbf{94.1\%}         &\textbf{96.9\%}         &84.3\%         &87.7\%          &97.8\%         &\textbf{98.4\%} \\
ResNet-101-C4             &82.1\%         &85.2\%              &92.4\%         &93.6\%             &84.9\%         &88.2\%              &\textbf{98.0\%}         &\textbf{98.4\%} \\
ResNet-101-DC5             &81.0\%         &84.5\%              &92.3\%         &94.0\%             &\textbf{85.5\%}         &\textbf{88.8\%}              &97.9\%         &\textbf{98.4\%} \\
ResNet-101-FPN             &79.5\%         &84.3\%              &92.9\%         &95.5\%             &82.0\%         &85.6\%              &97.5\%         &\textbf{98.4\%} \\
ResNeXt-101-FPN             &77.0\%         &82.0\%              &88.8\%         &92.4\%             &82.5\%         &86.2\%              &97.7\%         &98.2\% \\
\bottomrule
\end{tabular}}
\caption{Faster-RCNN Detection results.}
\label{tab:val_results}
\end{table}

%% file: sections/conclusion.tex
\section{Conclusion}
In this paper we present a novel rotated object detection framework for end-to-end solar panel detection and mapping. With our framework, we can directly predict the coordinates of individual solar panels and project them from image space back to geographical space. We believe that this ability to automatically map the coordinates of every solar panel in each array will provide the efficiency required to monitor the overall health of large-scale sites and help continue the rapid growth of the solar power capacity worldwide.

%% file: sections/appendix.tex
\newpage
\appendix

\section{Dataset Details}
\label{appendix:dataset}
\input{figures/usa_plot.tex}
We created a diverse dataset of aerial imagery collected from across North America. We collect 121 high resolution stitched orthomosaics from large scale solar farms using a fixed-wing aircraft with an RGB camera at roughly 2.5 cm ground sample distance (GSD). Of these orthomosaics, we reserve 10 specifically for testing.

\input{tables/data.tex}

\section{Implementation Details}
\label{appendix:implementation}

\subsection{Bounding Box Conversion}
We use the following equation given $\alpha$, the orientation parameter $\theta$ in radians to transform the compact oriented bounding box representation to the full set of vertex coordinates.

\begin{equation}
\label{eq:vertices}
\begin{aligned}
    x_1 &= x - \frac{w}{2} \cos(\alpha) + \frac{h}{2} \sin(\alpha), \ y_1 = y + \frac{w}{2} \cos(\alpha) + \frac{h}{2} \sin(\alpha) \\
    x_2 &= x - \frac{w}{2} \cos(\alpha) - \frac{h}{2} \sin(\alpha), \ y_2 = y + \frac{w}{2} \cos(\alpha) - \frac{h}{2} \sin(\alpha) \\
    x_3 &= x + \frac{w}{2} \cos(\alpha) - \frac{h}{2} \sin(\alpha), \ y_3 = y - \frac{w}{2} \cos(\alpha) - \frac{h}{2} \sin(\alpha) \\
    x_4 &= x + \frac{w}{2} \cos(\alpha) + \frac{h}{2} \sin(\alpha), \ y_4 = y - \frac{w}{2} \cos(\alpha) + \frac{h}{2} \sin(\alpha) \\
\end{aligned}
\end{equation}

\subsection{Training Hyperparameters}
We trained each baseline model with the hyperparameters in Table \ref{tab:hyperparameters} below. We train on 4 Tesla V100-SXM2-16GB GPUs which takes about 1 day to train for the non-FPN based models and about 4 days for the FPN models.

\input{tables/config.tex}

\subsection{Data Augmentation}
\label{section:augmentation}
During training we randomly augment the images and bounding boxes using the following methods:

\begin{itemize}
    \item Random Rotation described in \cite{RRPN} with probability $p=0.5$
    \item Color Jitter with brightness, contrast, saturation, and hue factors of $\sigma=0.1$ and probability $p=0.25$ 
    \item Contrast Limited Adaptive Histogram Equalization (CLAHE) \cite{zuiderveld1994contrast} with kernel size $k=8$ and probability $p=0.01$
    \item Random Blur with kernel size $k \in [3..7]$ and probability $p=0.01$
\end{itemize}

We find random rotation augmentation to be critical for accurate orientation prediction.

%% file: figures/usa_plot.tex
\begin{figure}[ht!]
\centering
\includegraphics[width=\textwidth]{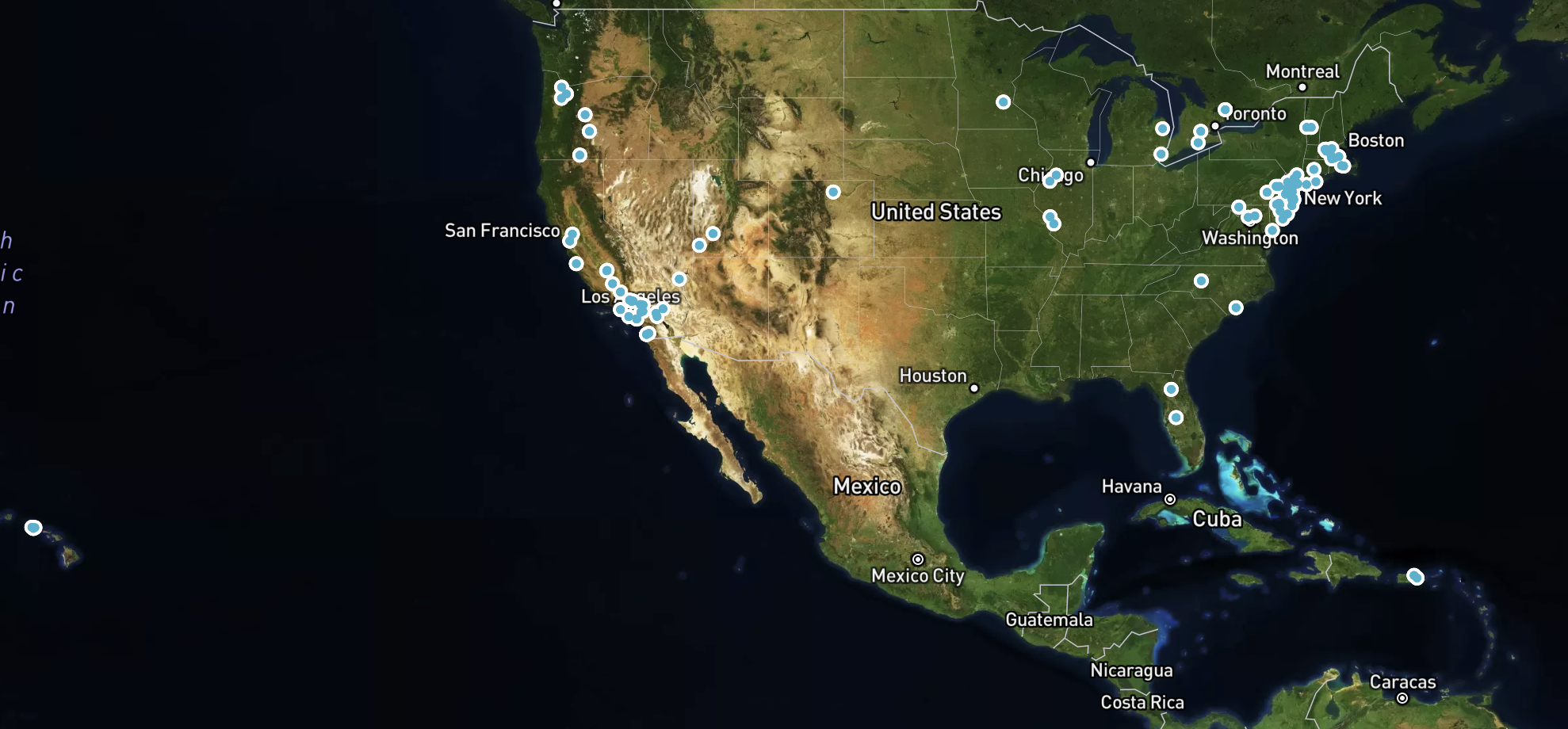}
\caption{USA Solar farm locations.}
\label{fig:usa-plot}
\end{figure}

%% file: tables/data.tex
\begin{table}[ht!]
\centering
\resizebox{\columnwidth}{!}{\begin{tabular}{@{}lcccccc@{}}
\toprule
\multicolumn{1}{c}{\textbf{Split}} &
\textbf{Background} &
\textbf{Foreground} &
\textbf{Total} &
\textbf{Sampled Background} &
\textbf{Sampled Foreground} &
\textbf{Total Sampled} \\ 
\midrule
Train         &57,323         &19,151          &76,474         &566         &795          &1,361 \\
Validation         &3,815         &1,133          &4,948         &88         &216          &304 \\
Test         &2,244         &192          &2,492         &-         &-          &- \\
\bottomrule
\end{tabular}}
\caption{Solar Panel Dataset.}
\label{tab:dataset}
\end{table}

%% file: tables/config.tex
\begin{table}[ht!]
\centering
\begin{tabular}{@{}ll@{}}
\toprule
\textbf{Hyperparameter} &
\textbf{Value} \\ 
\midrule
Optimizer      &Adam \\
Training Steps &$25,000$ \\
Learning Rate  &$1e-4$ \\
Linear Warmup  &$1e-3$ \\
Batch Size     &$64$ \\
Grad Norm Clip &$0.5$ \\
Angle Anchors  &$[-90^{\circ}, -60^{\circ}, -30^{\circ}, 0^{\circ}, 30^{\circ}, 60^{\circ}, 90^{\circ}]$ \\
Size Anchors   &$[32, 64, 128, 256, 512]$ \\
Aspect Ratio Anchors  &[1:2, 1:1, 2:1] \\
\bottomrule
\end{tabular}
\caption{Solar Panel Detection Hyperparameters.}
\label{tab:hyperparameters}
\end{table}

%% file: main.bbl
\begin{thebibliography}{15}
\providecommand{\natexlab}[1]{#1}
\providecommand{\url}[1]{\texttt{#1}}
\expandafter\ifx\csname urlstyle\endcsname\relax
  \providecommand{\doi}[1]{doi: #1}\else
  \providecommand{\doi}{doi: \begingroup \urlstyle{rm}\Url}\fi

\bibitem[Camilo et~al.(2017)Camilo, Wang, Collins, Bradbury, and Malof]{SegNet}
Joseph Camilo, Rui Wang, Leslie~M. Collins, Kyle Bradbury, and Jordan~M. Malof.
\newblock Application of a semantic segmentation convolutional neural network
  for accurate automatic detection and mapping of solar photovoltaic arrays in
  aerial imagery.
\newblock \emph{IEEE Applied Imagery Pattern Recognition (AIPR) Workshop},
  2017.

\bibitem[Dai et~al.(2017)Dai, Qi, Xiong, Li, Zhang, Hu, and Wei]{DC}
Jifeng Dai, Haozhi Qi, Yuwen Xiong, Yi~Li, Guodong Zhang, Han Hu, and Yichen
  Wei.
\newblock Deformable convolutional networks.
\newblock \emph{ICCV}, pp.\  764--773, 2017.

\bibitem[Deng et~al.(2009)Deng, Dong, Socher, Li, Li, and
  Fei-Fei]{imagenet_cvpr09}
Jia Deng, Wei Dong, Richard Socher, Li-Jia Li, Kai Li, and Li~Fei-Fei.
\newblock Imagenet: A large-scale hierarchical image database.
\newblock 2009.

\bibitem[Golovko et~al.(2021)Golovko, Kroshchanka, Mikhno, and
  Komar]{GolovkoEtAl}
Vladimir Golovko, Alexander Kroshchanka, Egor Mikhno, and Myroslav Komar.
\newblock Deep convolutional neural network for detection of solar panels.
\newblock \emph{Data-Centric Business and Applications}, pp.\  371--389, 2021.

\bibitem[Hou et~al.(2020)Hou, Wang, Hu, Yin, Huang, and Wu]{SolarNet}
Xin Hou, Biao Wang, Wanqi Hu, Lei Yin, Anbu Huang, and Haishan Wu.
\newblock Solarnet: A deep learning framework to map solar plants in china from
  satellite imagery.
\newblock \emph{ICLR}, 2020.

\bibitem[Khomiakov et~al.(2022)Khomiakov, Radzikowski, Schmidt, Sørensen,
  Andersen, Andersen, and Frellsen]{SolarDK}
Maxim Khomiakov, Julius Radzikowski, Car Schmidt, Mathias Sørensen, Mads
  Andersen, Michael Andersen, and Jes Frellsen.
\newblock Solardk: A high-resolution urban solar panel image classification and
  localization dataset.
\newblock \emph{NeurIPS}, 2022.

\bibitem[Lin et~al.(2017)Lin, Dollar, Girshick, He, Hariharan, and
  Belongie]{FPN}
Tsung-Yi Lin, Piotr Dollar, Ross Girshick, Kaiming He, Bharath Hariharan, and
  Serge Belongie.
\newblock Feature pyramid networks for object detection.
\newblock \emph{CVPR}, pp.\  2117--2125, 2017.

\bibitem[Ma et~al.(2018)Ma, Shao, Ye, Wang, Wang, Zheng, and Xue]{RRPN}
Jianqi Ma, Weiyuan Shao, Hao Ye, Li~Wang, Hong Wang, Yingbin Zheng, and
  Xiangyang Xue.
\newblock Arbitrary-oriented scene text detection via rotation proposals.
\newblock \emph{IEEE Transactions on Multimedia}, 20:\penalty0 3111--3122,
  2018.

\bibitem[Mujtaba \& Wani(2021)Mujtaba and Wani]{MujtabaAndWani}
Tahir Mujtaba and M.~Arif Wani.
\newblock Segmentation of satellite images of solar panels using fast deep
  learning model.
\newblock \emph{International Journal of Renewable Energy Research (IJRER)}, 11
  (1):\penalty0 31--45, 2021.

\bibitem[Parhar et~al.(2021)Parhar, Sawasaki, Nusaputra, Vergara, Todeschini,
  and Vahabi]{HyperionSolarNet}
Poonam Parhar, Ryan Sawasaki, Nathan Nusaputra, Felipe Vergara, Alberto
  Todeschini, and Hossein Vahabi.
\newblock Hyperionsolarnet solar panel detection from aerial images.
\newblock \emph{NeurIPS}, 2021.

\bibitem[Rapier(2020)]{renewables}
Robert Rapier.
\newblock Renewable energy growth continues at a blistering pace, 2020.
\newblock URL
  \url{https://www.forbes.com/sites/rrapier/2020/08/02/renewable-energy-growth-continues-at-a-blistering-pace/?sh=2fcf9f3276b6}.
\newblock Forbes Magazine.

\bibitem[Ren et~al.(2015)Ren, He, Girshick, and Sun]{FasterRCNN}
Shaoqing Ren, Kaiming He, Ross Girshick, and Jian Sun.
\newblock Faster r-cnn: Towards real-time object detection with region proposal
  networks.
\newblock \emph{NeurIPS}, 2015.

\bibitem[Yu et~al.(2018)Yu, Wang, Majumdar, and Rajagopal]{DeepSolar}
Jiafan Yu, Zhecheng Wang, Arun Majumdar, and Ram Rajagopal.
\newblock Deepsolar: A machine learning framework to efficiently construct a
  solar deployment database in the united states.
\newblock \emph{Joule}, 2:\penalty0 2605–2617, 2018.

\bibitem[Zhuang et~al.(2020)Zhuang, Zhang, and Wang]{CrossNets}
Li~Zhuang, Zijun Zhang, and Long Wang.
\newblock The automatic segmentation of residential solar panels based on
  satellite images: A cross learning driven u-net method.
\newblock \emph{Applied Soft Computing Journal}, 92, 2020.

\bibitem[Zuiderveld(1994)]{zuiderveld1994contrast}
Karel Zuiderveld.
\newblock Contrast limited adaptive histogram equalization.
\newblock \emph{Graphics gems}, pp.\  474--485, 1994.

\end{thebibliography}
